\documentclass{article}

\PassOptionsToPackage{numbers, compress}{natbib}

\usepackage[preprint]{neurips_2025}

\usepackage[utf8]{inputenc} 
\usepackage[T1]{fontenc}    
\usepackage{hyperref}       
\usepackage{url}            
\usepackage{booktabs}       
\usepackage{amsfonts}       
\usepackage[table]{xcolor}  
\usepackage{nicefrac}       
\usepackage{microtype}      
\usepackage{xcolor}         
\usepackage{graphicx}
\usepackage{multirow}
\usepackage{pifont}
\usepackage{amssymb}
\usepackage{chngcntr}
\counterwithout{table}{section}
\usepackage[most]{tcolorbox}
\usepackage{longtable}

\usepackage{makecell}
\usepackage{threeparttable}
\usepackage{array}
\newcolumntype{P}[1]{>{\centering\arraybackslash}p{#1}}

\newcommand{\eg}{\textit{e.g.}, }

\newcommand{\wrt}{\textit{w.r.t. }}

\title{Are Vision Language Models Ready for Clinical Diagnosis? A 3D Medical Benchmark for Tumor-centric Visual Question Answering}

\author{
\textbf{Yixiong Chen}$^{1}$,
\textbf{Wenjie Xiao}$^{1}$,
\textbf{Pedro R. A. S. Bassi}$^{1,2,4}$,
\textbf{Xinze Zhou}$^{1}$, \\
\textbf{Sezgin Er}$^{3}$,
\textbf{Ibrahim Ethem Hamamci}$^{3}$,
\textbf{Zongwei Zhou}$^{1}$,
\textbf{Alan Yuille}$^{1}$ \\
$^{1}$Johns Hopkins University \quad
$^{2}$University of Bologna \quad
$^{3}$Istanbul Medipol University \quad \\
$^{4}$Center for Biomolecular Nanotechnologies, Istituto Italiano di Tecnologia \\
\texttt{ayuille1@jhu.edu}
}

\begin{document}

\maketitle

\begin{abstract}
Vision-Language Models (VLMs) have shown promise in various 2D visual tasks, yet their readiness for 3D clinical diagnosis remains unclear due to stringent demands for recognition precision, reasoning ability, and domain knowledge. To systematically evaluate these dimensions, we present DeepTumorVQA, a diagnostic visual question answering (VQA) benchmark targeting abdominal tumors in CT scans. It comprises 9,262 CT volumes (3.7M slices) from 17 public datasets, with 395K expert-level questions spanning four categories: \emph{Recognition}, \emph{Measurement}, \emph{Visual Reasoning}, and \emph{Medical Reasoning}. 
DeepTumorVQA introduces unique challenges, including small tumor detection and clinical reasoning across 3D anatomy.
Benchmarking four advanced VLMs (RadFM, M3D, Merlin, CT-CHAT), we find current models perform adequately on measurement tasks but struggle with lesion recognition and reasoning, and are still not meeting clinical needs. Two key insights emerge: (1) large-scale multimodal pretraining plays a crucial role in DeepTumorVQA testing performance, making RadFM stand out among all VLMs. (2) Our dataset exposes critical differences in VLM components, where proper image preprocessing and design of vision modules significantly affect 3D perception.
To facilitate medical multimodal research, we have released DeepTumorVQA as a rigorous benchmark: \href{https://github.com/Schuture/DeepTumorVQA}{https://github.com/Schuture/DeepTumorVQA}.
\end{abstract}

\section{Introduction}
\label{sect:intro}

Vision-language models (VLMs) \citep{zhang2024vision} have achieved impressive performance across general visual reasoning tasks. However, applying them to medical imaging introduces significantly more stringent requirements, due to the high-stakes nature of clinical decision-making. Existing medical VLMs \citep{wu2023towards,hamamci2024ct2rep,bai2024m3d,blankemeier2024merlin} have typically been evaluated on simplified or exploratory benchmarks that do not reflect real-world clinical complexity. This raises a critical question: \textit{Are 3D medical VLMs precise and intelligent enough for clinical diagnosis?} 
Clinical diagnosis refers to the judgment about the nature of a patient’s disease, made by imaging studies in the context of this work.
To address this, there is a pressing need for a high-quality and diagnostically meaningful benchmark that enables rigorous evaluation of state-of-the-art (SOTA) models in realistic clinical contexts.

A number of medical VQA benchmarks~\citep{lin2023medical,hartsock2024vision} have been proposed to evaluate the capabilities of VLMs. However, they suffer from five limitations that hinder their utility as standardized benchmarks:
\textbf{First}, \emph{limited scale and diversity}. Due to the high cost and time required for expert annotation, most clinical datasets remain small in scale and lack diversity (\eg VQA-Rad~\citep{lau2018dataset}, VQA-Med~\citep{ben2021overview}, Open-I~\citep{demner2016preparing}, EndoVis 2017~\citep{allan20192017}).
\textbf{Second}, \emph{reliance on 2D and web-sourced images}. Many recent large-scale datasets, including SLAKE~\citep{liu2021slake}, PMC-VQA~\citep{zhang2023pmc}, OmniMedVQA~\citep{hu2024omnimedvqa}, and PathVQA~\citep{he2020pathvqa}, are constructed using 2D images from public websites or scientific publications. They do not adequately reflect the 3D volumetric nature of clinical imaging.
\textbf{Third}, \emph{lack of consistent and reliable evaluation metrics}. Automated metrics such as BLEU and ROUGE are not well-suited for evaluating short, factual medical answers, as they often fail to capture semantic correctness~\citep{lin2023medical}. While human evaluation~\citep{kovaleva2020towards} aligns more closely with clinical judgment, it is costly and difficult to reproduce.
\textbf{Fourth}, \emph{oversimplified questions}. Existing datasets often include experimental or toy questions (\eg organ, phase, or plane recognition~\citep{bai2024m3d}). However, real-world clinical questions frequently require measurement and reasoning with anatomical knowledge and clinical context.
\textbf{Fifth}, \emph{limited accessibility}. Some datasets are based on private institutional data, which restricts broad usage and reproducibility in the research community.
To date, no existing medical VQA dataset integrates large-scale, multi-source 3D imaging data with high-quality expert annotations and clinically structured question hierarchies into a unified and accessible benchmark.

\begin{figure}
    \centering
    \includegraphics[width=1.0\linewidth]{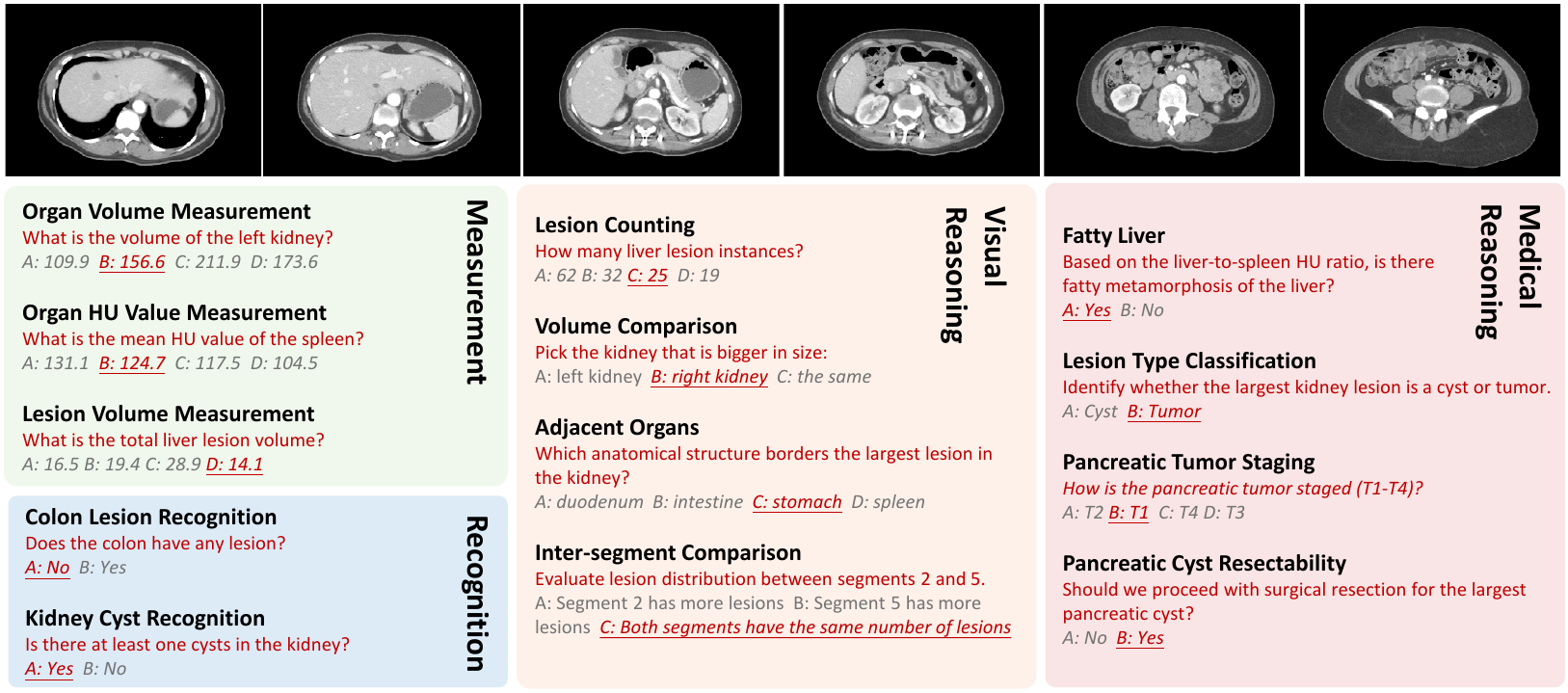}
    \caption{Overview of tasks in the DeepTumorVQA benchmark. The dataset covers four core clinical question types, totaling 29 subtypes. Tasks include numerical quantification (\eg organ volume, Hounsfield Unit (HU) value), lesion recognition, spatial reasoning (\eg comparisons, adjacency), and high-level clinical diagnosis (\eg tumor staging, resectability). Each question is paired with image evidence and formatted for either multiple-choice or free-text answer prediction, enabling evaluation of both perceptual and diagnostic reasoning in VLMs.}
    \vspace{-0.3cm}
    \label{fig:task_dem}
\end{figure}

To bridge these gaps, we introduce \textbf{DeepTumorVQA} (Fig. \ref{fig:task_dem}), a comprehensive dataset for evaluating VLMs in abdominal CT-based clinical diagnostics. DeepTumorVQA comprises 9,262 CT volumes (3.7M slices) derived from 17 public datasets and 88 centers.
Over 20 board-certified radiologists participated in the annotation and all questions are generated using templates from patterns found in structured radiology reports and medical literature, ensuring clinical relevance. DeepTumorVQA comprises 395K question-answer pairs covering four hierarchical diagnostic tasks: \emph{Recognition}, \emph{Measurement}, \emph{Visual Reasoning}, and \emph{Medical Reasoning}. 
The former two types require models to precisely perceive organs and lesions.
Built upon them, the latter two require models to intelligently reason about anatomical structures and apply external medical knowledge.
The dataset mirrors the diagnostic reasoning hierarchy used by radiologists.

Through extensive benchmarking experiments using four SOTA VLMs—RadFM \citep{wu2023towards}, M3D \citep{bai2024m3d}, Merlin \citep{blankemeier2024merlin}, and CT-CHAT \citep{hamamci2024developing}—we provide detailed analyses that expose fundamental strengths and weaknesses of existing approaches. The results show that SOTA VLMs are better at large objects like organs, but struggle significantly with identifying small lesions and performing reasoning tasks that involve them. 
Our in-depth analysis also reveals the impact of basic visual tasks on the reasoning tasks, as well as the relationship between lesion characteristics and recognition performance.

Our contributions are summarized as follows: (1) We release DeepTumorVQA, the first large-scale 3D VQA benchmark for tumor diagnosis with expert annotations and question hierarchies. (2) We present a comprehensive empirical analysis of VLMs, revealing key challenges in lesion recognition and reasoning. (3) We provide open-source data, code, and tools, and commit to maintaining the benchmark via recurring challenges.

\section{Related Work}
\label{sect:related_work}

\textbf{Medical Visual Question Answering.} VQA has become an important benchmark task for evaluating multimodal clinical systems. Early medical VQA datasets such as VQA-RAD~\citep{lau2018dataset} and VQA-Med 2018--2020~\citep{ben2021overview} featured small-scale 2D image collections with a limited range of question types, often relying on templates or handcrafted QA pairs. Subsequently, more diverse datasets like PathVQA~\citep{he2020pathvqa}, SLAKE~\citep{liu2021slake}, and RadVisDial~\citep{kovaleva2020towards} introduced pathology slides, structured medical knowledge, and dialog-style multi-turn QA, broadening the scope beyond simple abnormality detection. Recently, larger-scale benchmarks such as PMC-VQA~\citep{zhang2023pmc}, and OmniMedVQA~\citep{hu2024omnimedvqa} incorporate richer question types, hierarchical QA structures, and answers grounded in dense clinical reports. These datasets have shifted the field’s emphasis from classification to explanation, reasoning, and domain adaptation. Notably, as public datasets like RadGenome Chest-CT~\citep{zhang2024radgenome}, RadGenome Brain-MRI~\citep{lei2024autorg}, and AMOS~\citep{ji2022amos} have expanded, the feasibility of 3D medical VQA has improved significantly, enabling the creation of volumetric benchmarks requiring spatial reasoning and multi-slice integration~\citep{hamamci2024developing}. This transition from static 2D diagnosis to rich, multi-view 3D reasoning reflects the evolution of the task’s complexity and its alignment with real-world clinical scenarios.

\textbf{Medical Vision-Language Models.} VLMs designed for medical imaging tasks have undergone significant architectural and methodological evolution. Earlier systems largely used ResNet-based \citep{he2016deep} image encoders paired with LSTM or Transformer-based text encoders~\citep{sharma2021medfusenet,abacha2018nlm,peng2018umass,ren2020cgmvqa}. Recent models have transitioned to Vision Transformer (ViT) backbones~\citep{hamamci2024generatect}, which better preserve spatial and contextual information, and allow for more expressive visual representations. Pretraining objectives have shifted from contrastive learning (CLIP-style)~\citep{wang2022medclip,zhang2023large} to encoder-decoder paradigms, where image features are passed into large language decoders for autoregressive medical text generation ~\citep{chen2025coca}. Concurrently, models like Med-PaLM~\citep{tu2024towards}, LLaVA-Med~\citep{li2023llava}, Med-Gemini~\citep{yang2024advancing}, and RadFM~\citep{wu2023towards} began to scale both in terms of language model size and the diversity of medical tasks they support. Another recent trend is the support for 3D inputs, where ResNets/ViTs are adapted to volumetric data~\citep{bai2024m3d,blankemeier2024merlin} and 3D image-text pretraining. Additionally, the pretraining corpora have evolved to include multiple clinical data sources—reports, textbooks, biomedical QA pairs, and PACS metadata—making modern medical VLMs increasingly robust and generalizable across domains.

\section{DeepTumorVQA Dataset}
\label{sect:dataset}

\subsection{Overview}

The design of \textbf{DeepTumorVQA} is inspired by the compositional reasoning framework in CLEVR~\citep{johnson2017clevr}, adapted to the clinical context of diagnostic decision-making in abdominal CT. Our goal is to build a dataset that reflects real-world diagnostic needs while exposing the performance boundaries of VLMs under varying levels of task complexity. DeepTumorVQA comprises basic and compositional question types, ranging from simple recognition and measurement to sophisticated visual and clinical reasoning, thus enabling detailed analysis of VLM behavior and limitations.

To overcome the limitations highlighted in Section~\ref{sect:intro}, we construct a large-scale benchmark featuring:
(1) \textbf{High data volume and diversity}: We curate 3D CT scans from 17 public datasets, encompassing over 9,000 volumes and millions of slices.
(2) \textbf{Volumetric 3D supervision}: Unlike most prior benchmarks limited to 2D images, our dataset operates on full CT volumes, aligning with clinical diagnostic practice.
(3) \textbf{Standardized evaluation metrics}: To ensure reproducibility and clinical relevance, we use task-specific metrics: accuracy for multiple-choice questions, exact match for free-text categorical answers, and mean relative accuracy (MRA) \citep{yang2024thinking} for quantitative numerical prediction.
(4) \textbf{Clinical question design}: These question types align with key steps in radiological workflows, where clinicians must not only perceive features but also reason about their diagnostic significance. Importantly, reasoning questions are systematically constructed by composing functions over outputs from the recognition and measurement stages. This can ensure a dependency structure among questions, acting as a smart way to enforce multi-step reasoning.

\begin{figure}
    \centering
    \includegraphics[width=1.0\linewidth]{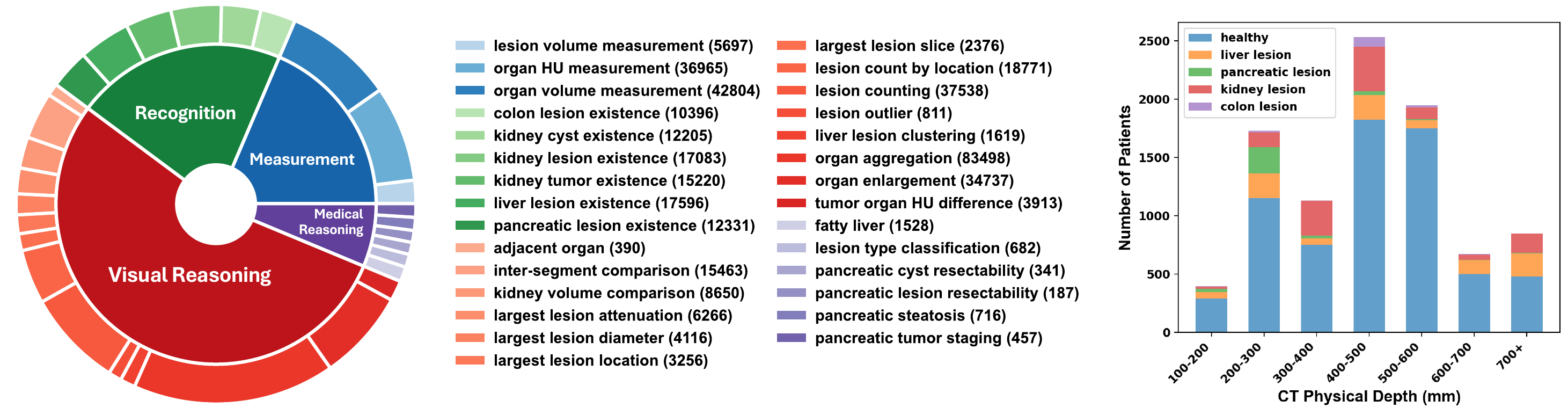}
    \caption{Statistics of DeepTumorVQA. Left: the distribution of QA pairs for tasks across four main types. Right: distribution of CT volumes \wrt CT physical depth (z-axis) and patient types.}
    \label{fig:stat}
\end{figure}

The dataset contains 355,962 training QA pairs from 8,334 CT and 39,650 testing QA pairs from 928 CT. Its statistics of tasks and CT samples are shown in Fig. \ref{fig:stat}.

\subsection{Dataset Construction}
\label{sect:data_construct}

\paragraph{Data Collection.}  
We compile 9,262 abdominal CT volumes from 17 public datasets (Tab.~\ref{tab:data_overview}), encompassing diverse acquisition protocols, scanners, and patient populations from 88 centers to ensure robust coverage of organs and pathologies.

\begin{table*}[t]
\centering
\scriptsize
\begin{threeparttable}
\caption{Overview of public abdominal CT datasets that are collected in DeepTumorVQA. Our reported number of CT volumes may differ from the original publications, as some CT volumes are reserved for further validation purposes. The number of CT volumes in DeepTumorVQA is lower than the sum of datasets 1--17 due to the removal of duplicated samples.}
\label{tab:data_overview}
\setlength{\tabcolsep}{4pt}
\begin{tabular}{
    p{0.25\linewidth}
    P{0.1\linewidth}
    P{0.1\linewidth}
    |
    p{0.25\linewidth}
    P{0.1\linewidth}
    P{0.1\linewidth}
}
    \toprule
    dataset (year) [source]
        & \makecell{\# of volumes}
        & \makecell{\# of centers}
    & dataset (year) [source]
        & \makecell{\# of volumes}
        & \makecell{\# of centers}
    \\
    \midrule
    1.\ CHAOS \citeyearpar{valindria2018multi}
    [\href{https://chaos.grand-challenge.org/Download/}{link}]
        & 20 & 1
    & 2.\ Pancreas-CT \citeyearpar{roth2015deeporgan}
    [\href{https://academictorrents.com/details/80ecfefcabede760cdbdf63e38986501f7becd49}{link}]
        & 42 & 1
    \\
    3.\ BTCV \citeyearpar{landman2015miccai}
    [\href{https://www.synapse.org/#!Synapse:syn3193805/wiki/89480}{link}]
        & 47 & 1
    & 4.\ LiTS \citeyearpar{bilic2019liver}
    [\href{https://competitions.codalab.org/competitions/17094}{link}]
        & 131 & 7
    \\
    5.\ CT-ORG \citeyearpar{rister2020ct}
    [\href{https://wiki.cancerimagingarchive.net/pages/viewpage.action?pageId=61080890#61080890cd4d3499fa294f489bf1ea261184fd24}{link}]
        & 140 & 8
    & 6.\ WORD \citeyearpar{luo2021word}
    [\href{https://github.com/HiLab-git/WORD}{link}]
        & 120 & 1
    \\
    7.\ AMOS22 \citeyearpar{ji2022amos}
    [\href{https://amos22.grand-challenge.org}{link}]
        & 200 & 2
    & 8.\ KiTS \citeyearpar{heller2020international}
    [\href{https://kits-challenge.org/kits23/}{link}]
        & 489 & 1
    \\
    9--14.\ MSD CT Tasks \citeyearpar{antonelli2021medical}
    [\href{https://decathlon-10.grand-challenge.org/}{link}]
        & 945 & 1
    & 15.\ AbdomenCT-1K \citeyearpar{ma2021abdomenct}
    [\href{https://github.com/JunMa11/AbdomenCT-1K}{link}]
        & 1,050 & 12
    \\
    16.\ FLARE'23 \citeyearpar{ma2022fast}
    [\href{https://codalab.lisn.upsaclay.fr/competitions/12239}{link}]
        & 4,100 & 30
    & 17.\ Trauma Detect.\ \citeyearpar{rsna-2023-abdominal-trauma-detection}
    [\href{https://www.rsna.org/education/ai-resources-and-training/ai-image-challenge/abdominal-trauma-detection-ai-challenge}{link}]
        & 4,711 & 23
    \\
    \bottomrule
\end{tabular}
\end{threeparttable}
\end{table*}

To provide high-quality annotations, 23 board-certified radiologists manually annotated 7,629 lesions over six months, including 3,067 liver, 4,078 kidney, 351 pancreatic, and 131 colon lesions. Kidney tumors and cysts were labeled when distinguishable; ambiguous cases were marked as non-specific lesions. All annotations were performed in 3D and double-checked for consensus.

\begin{figure}
    \centering
    \includegraphics[width=1.0\linewidth]{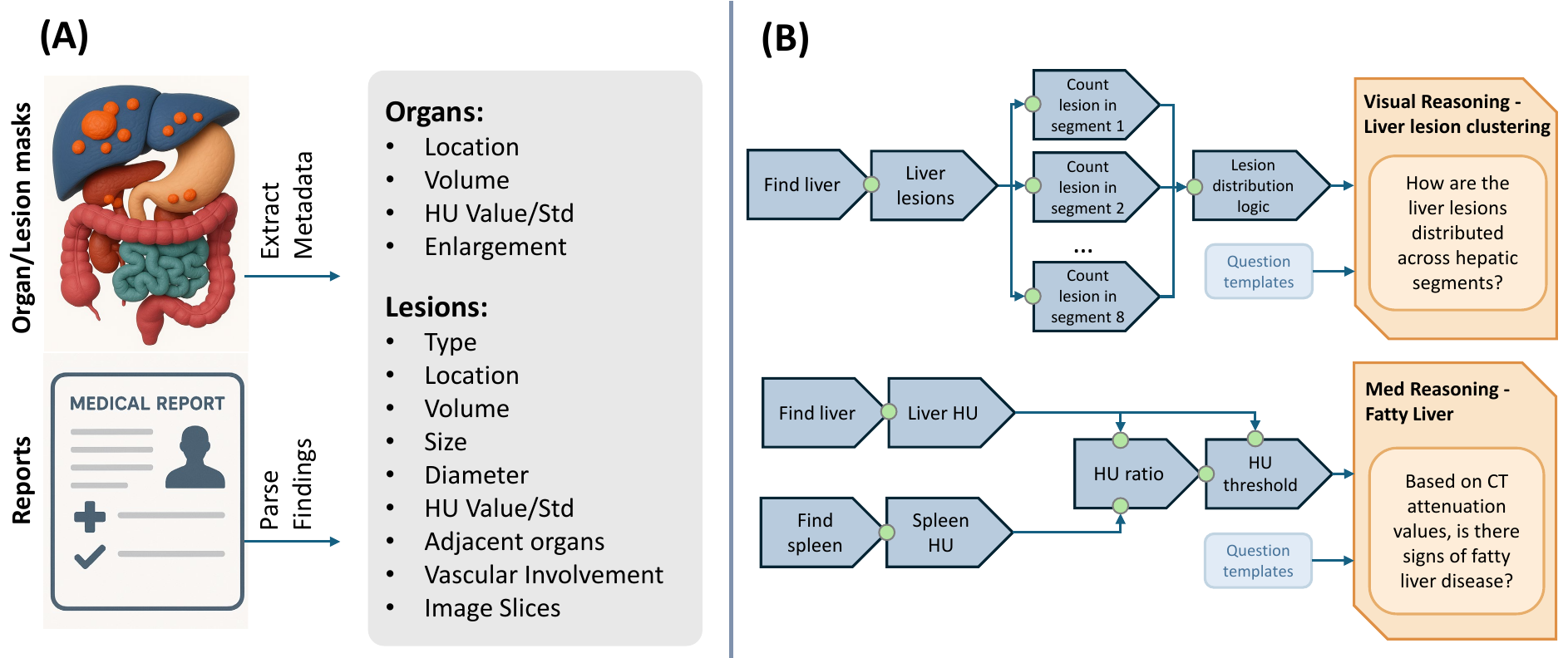}
    \caption{
    Overview of question construction in the DeepTumorVQA dataset. \textbf{(A)} Structured metadata is extracted from organ and lesion segmentation masks (\eg location, volume, HU value, enlargement) and parsed radiology reports (\eg lesion type, adjacent organs, vascular involvement). \textbf{(B)} These metadata are used to define modular logic programs for different diagnostic question types. For example, liver segment-level lesion counts are used to construct distribution-based visual reasoning questions. Each program maps to one of four task types and 29 subtypes, and is rendered into natural language using predefined question templates.
    }
    \label{fig:dataset_construction}
\end{figure}

\paragraph{Question Generation.}  
We adopt a modular, rule-based approach inspired by CLEVR's functional program generation~\citep{johnson2017clevr} to construct question-answer (QA) pairs (Figure~\ref{fig:dataset_construction}) in the context of clinical diagnosis. 
In this work, we define \textbf{clinical diagnosis} as the process of interpreting radiological images to identify and assess the clinical implications of abnormalities, especially tumors. Based on this, we define four types of diagnostic tasks with increasing complexity:

\textbf{Measurement (3 subtypes):} Numerical assessments like organ volume and HU value.

\textbf{Recognition (6 subtypes):} Recognize lesions like tumors and cysts.

\textbf{Visual Reasoning (14 subtypes):} Compositional logic-based tasks including spatial comparisons (\eg “Which segment contains more lesions?”), counting and localization of lesions, and lesion-organ relationship (\eg ``Are there adjacent organs for a specific tumor?").

\textbf{Medical Reasoning (6 subtypes):} Clinical inference tasks requiring external knowledge from clinical literature, such as fatty liver diagnosis \citep{zeb2012computed}, kidney lesion diagnosis \citep{agochukwu2017differentiating}, pancreatic steatosis diagnosis \citep{guneyli2022computed}, pancreatic cyst resectability \citep{hopkins_pancreatic_cyst_2022}, and pancreatic tumor staging \citep{bassi2025radgpt}.

Each question is generated via a structured program and templated prompts. For example, reasoning about hepatic lesion distribution is computed from segment-level tumor burden. 
In addition to multi-choice questions, we also hide the choices to serve as free-text questions in the DeepTumorVQA dataset. In this case, the model must predict text-form answers.
Our dataset generation heavily relies on the radiologists' annotation of organ/lesion masks. Careful design of the question generation pipeline is crucial for correctness. We summarize the specific metadata extraction logistics and the full question definition/generation details in the Appendix \ref{sect:appendix_A} and \ref{sect:appendix_B}. Although the current DeepTumorVQA contains multiple lesions, we would expand our annotation to other anatomies in future versions.

\section{Evaluation on DeepTumorVQA Benchmark}
\label{sect:eval}

\subsection{Details of VLM Evaluation}
\label{sect:vlms}

To evaluate the capabilities of current VLMs in solving volumetric medical VQA tasks, we benchmark four representative models: RadFM \citep{wu2023towards}, M3D \citep{bai2024m3d}, Merlin \citep{blankemeier2024merlin}, and CT-CHAT \citep{hamamci2024developing}. Each model adopts a different architectural design and training strategy to integrate 3D visual information with language modeling, as summarized in Tab. \ref{tab:train_detail}.

\begin{table*}[ht]
\centering
\scriptsize
\caption{Model architectures and training settings for four benchmarked VLMs. We use the original code base of the methods and follow their training hyperparameters for VQA tasks.}
\begin{tabular}{lcccc}
\toprule
\textbf{Component} & \textbf{RadFM} & \textbf{M3D (LLaMA2 / Phi-3)} & \textbf{Merlin} & \textbf{CT-CHAT} \\
\midrule
Vision Encoder & 3D ViT & 3D ViT & 3D ResNet & CT ViT \\
Input image size & [256,256,64] & [256,256,32] & [224,224,160] & [300,300,600] \\
3D CT spacing & direct resize & direct resize & [1.5mm, 1.5mm, 3mm] & [1.5mm, 1.5mm, 1.5mm] \\
LLM Decoder & LLaMA2-13B & LLaMA2-7B / Phi-3-4B & RadLLaMA-7B & LLaMA3.1-7B \\
Projector & Perceiver Resampler & 3D Pooling + 2-layer MLP & 1-layer FC & CoCa Attentional Pooling \\
Visual tokens/image & 32 & 256 & 1 & 256 \\
Pretraining Data & 16M 2D+3D multimodal & 120K 3D CT & 14K 3D Abdomen CT & 50K 3D Lung CT \\
LLM tuning & full & LoRA (r=16) & LoRA (r=128) & LoRA (r=128) \\
Projector tuning & \ding{51} & \ding{51} & \ding{51} & \ding{51} \\
Vision tuning & \ding{55} & \ding{55} & \ding{55} & \ding{55} \\
Learning rate & 5e-6 & 5e-5 & 1e-4 & 2e-5 \\
\bottomrule
\end{tabular}
\label{tab:train_detail}
\end{table*}

Each model varies in its architectural design and training procedure. RadFM leverages a pre-trained LLaMA2-13B \citep{touvron2023llama} with a perceiver resampler \citep{alayrac2022flamingo} to fuse 3D features, and fine-tunes both the LLM decoder and projector. M3D supports two LLM backbones (LLaMA2 and Phi-3 \citep{abdin2024phi}) and adopts a spatial pooling perceiver module to aggregate 3D volume features. In contrast, Merlin uses a simpler architecture with a ResNet-based visual encoder \citep{hara3dcnns} and a single-layer linear projector. Finally, CT-CHAT adopts a ViT-based encoder tailored for CT \citep{hamamci2024generatect} and employs a CoCa attentional pooling \citep{yu2022coca}. All the latter three models are fine-tuned with LoRA. We summarize the training hyperparameters and computation costs in Appendix \ref{sect:appendix_C}.

\subsection{Analysis of Benchmarking Results}
\label{sect:benchmark}

Table~\ref{tab:vqa_full_comparison} reports performance across five VLMs under both multi-choice and free-text settings. We analyze model behaviors along three axes: input format, diagnostic task types, and architecture.

\paragraph{1. Multi-choice questions yield higher accuracy than free-text.}
Four of five models perform better in the multi-choice setting, where candidate options provide inductive constraints. For example, in \textit{lesion counting}, models generate plausible answers with choices, but default to zero in free-text, indicating weak numeracy, especially for small structures.
However, this advantage diminishes for Yes/No-style questions in \textit{recognition} and binary reasoning tasks (\eg \textit{fatty liver}, \textit{pancreatic steatosis}), where free-text matches or even slightly exceeds multi-choice. This may stem from pretraining on open-ended generation, which favors categorical outputs.

\begin{table*}[t]
\centering
\scriptsize
\setlength{\tabcolsep}{2pt}
\caption{Performance for five VLMs under multi-choice and free-text settings. Subtypes marked with $^{*}$ indicate free-text numerical answers evaluated using MRA, higher is better. Meas. = Measurement, Recog. = Recognition, Vis. Rsn. = Visual Reasoning, Med. Rsn. = Medical Reasoning.}
\begin{tabular}{l l | c c c c c c | c c c c c}
\toprule
\multirow{2}{*}{\textbf{Type}} & \multirow{2}{*}{\textbf{Subtype}} & \multicolumn{6}{c|}{\textbf{Multi-choice}} & \multicolumn{5}{c}{\textbf{Free-text}} \\
 &  &  \textcolor{gray}{Rand} & Merlin & M3D-L2 & M3D-P3 & CT-CHAT & RadFM & Merlin & M3D-L2 & M3D-P3 & CT-CHAT & RadFM \\
\midrule
\multirow{3}{*}{\rotatebox{90}{Meas.}} 
& lesion volume measurement$^{*}$ & \textcolor{gray}{0.250} & 0.253 & 0.815 & 0.825 & 0.833 & 0.815 & 0.079 & 0.085 & 0.079 & 0.075 & 0.112 \\
& organ HU measurement$^{*}$ & \textcolor{gray}{0.250} & 0.254 & 0.638 & 0.640 & 0.637 & 0.647 & 0.487 & 0.490 & 0.491 & 0.513 & 0.608 \\
& organ volume measurement$^{*}$ & \textcolor{gray}{0.250} & 0.262 & 0.747 & 0.754 & 0.750 & 0.755 & 0.526 & 0.535 & 0.528 & 0.549 & 0.583 \\
\multicolumn{2}{c|}{\textbf{Average}} & \textcolor{gray}{0.250} & 0.256 & 0.733 & \textbf{0.740} & \textbf{0.740} & 0.739 & 0.364 & 0.370 & 0.366 & 0.379 & \textbf{0.434} \\
\midrule
\multirow{6}{*}{\rotatebox{90}{Recog.}} 
& colon lesion existence & \textcolor{gray}{0.500} & 0.859 & 0.859 & 0.859 & 0.859 & 0.856 & 0.859 & 0.859 & 0.859 & 0.859 & 0.893 \\
& kidney cyst existence & \textcolor{gray}{0.500} & 0.797 & 0.797 & 0.797 & 0.797 & 0.861 & 0.797 & 0.797 & 0.797 & 0.797 & 0.864 \\
& kidney lesion existence & \textcolor{gray}{0.500} & 0.495 & 0.510 & 0.501 & 0.514 & 0.668 & 0.511 & 0.515 & 0.490 & 0.507 & 0.692 \\
& kidney tumor existence & \textcolor{gray}{0.500} & 0.564 & 0.574 & 0.574 & 0.574 & 0.886 & 0.574 & 0.574 & 0.574 & 0.574 & 0.890 \\
& liver lesion existence & \textcolor{gray}{0.500} & 0.535 & 0.524 & 0.517 & 0.524 & 0.652 & 0.524 & 0.524 & 0.524 & 0.524 & 0.662 \\
& pancreatic lesion existence & \textcolor{gray}{0.500} & 0.718 & 0.718 & 0.718 & 0.718 & 0.810 & 0.718 & 0.718 & 0.718 & 0.718 & 0.871 \\
\multicolumn{2}{c|}{\textbf{Average}} & \textcolor{gray}{0.500} & 0.661 & 0.664 & 0.661 & 0.664 & \textbf{0.789} & 0.664 & 0.665 & 0.660 & 0.663 & \textbf{0.812} \\
\midrule
\multirow{14}{*}{\rotatebox{90}{Vis. Rsn.}} 
& adjacent organ & \textcolor{gray}{0.333} & 0.217 & 0.565 & 0.609 & 0.609 & 0.609 & 0.174 & 0.174 & 0.304 & 0.304 & 0.435 \\
& inter-segment comparison & \textcolor{gray}{0.333} & 0.470 & 0.567 & 0.576 & 0.572 & 0.591 & 0.577 & 0.561 & 0.592 & 0.589 & 0.456 \\
& kidney volume comparison & \textcolor{gray}{0.333} & 0.347 & 0.370 & 0.364 & 0.372 & 0.386 & 0.350 & 0.370 & 0.356 & 0.370 & 0.386 \\
& largest lesion attenuation & \textcolor{gray}{0.333} & 0.317 & 0.541 & 0.539 & 0.544 & 0.555 & 0.526 & 0.544 & 0.548 & 0.542 & 0.521 \\
& largest lesion diameter$^{*}$ & \textcolor{gray}{0.250} & 0.263 & 0.778 & 0.783 & 0.781 & 0.766 & 0.182 & 0.209 & 0.233 & 0.269 & 0.232 \\
& largest lesion location & \textcolor{gray}{0.392} & 0.307 & 0.310 & 0.310 & 0.340 & 0.340 & 0.359 & 0.353 & 0.337 & 0.353 & 0.334 \\
& largest lesion slice$^{*}$ & \textcolor{gray}{0.250} & 0.241 & 0.672 & 0.684 & 0.672 & 0.664 & 0.524 & 0.533 & 0.510 & 0.513 & 0.672 \\
& lesion count by location$^{*}$ & \textcolor{gray}{0.250} & 0.583 & 0.861 & 0.860 & 0.862 & 0.861 & 0.534 & 0.534 & 0.534 & 0.534 & 0.506 \\
& lesion counting$^{*}$ & \textcolor{gray}{0.328} & 0.455 & 0.781 & 0.784 & 0.796 & 0.790 & 0.000 & 0.000 & 0.000 & 0.000 & 0.001 \\
& lesion outlier & \textcolor{gray}{0.500} & 0.521 & 0.507 & 0.549 & 0.451 & 0.493 & 0.451 & 0.535 & 0.535 & 0.577 & 0.521 \\
& liver lesion clustering & \textcolor{gray}{0.333} & 0.331 & 0.438 & 0.475 & 0.463 & 0.469 & 0.388 & 0.469 & 0.469 & 0.431 & 0.513 \\
& organ aggregation$^{*}$ & \textcolor{gray}{0.250} & 0.257 & 0.660 & 0.667 & 0.655 & 0.661 & 0.577 & 0.569 & 0.586 & 0.574 & 0.621 \\
& organ enlargement & \textcolor{gray}{0.500} & 0.736 & 0.736 & 0.736 & 0.736 & 0.746 & 0.736 & 0.736 & 0.736 & 0.736 & 0.759 \\
& tumor organ HU difference$^{*}$ & \textcolor{gray}{0.305} & 0.296 & 0.836 & 0.839 & 0.821 & 0.821 & 0.113 & 0.122 & 0.139 & 0.197 & 0.189 \\
\multicolumn{2}{c|}{\textbf{Average}} & \textcolor{gray}{0.335} & 0.382 & 0.616 & \textbf{0.627} & 0.620 & 0.625 & 0.392 & 0.408 & 0.420 & 0.428 & \textbf{0.439} \\
\midrule
\multirow{6}{*}{\rotatebox{90}{Med. Rsn.}} 
& fatty liver & \textcolor{gray}{0.333} & 0.318 & 0.461 & 0.455 & 0.481 & 0.481 & 0.481 & 0.481 & 0.396 & 0.487 & 0.578 \\
& lesion type classification & \textcolor{gray}{0.500} & 0.865 & 0.865 & 0.865 & 0.865 & 0.865 & 0.865 & 0.865 & 0.865 & 0.865 & 0.851 \\
& pancreatic cyst resectability & \textcolor{gray}{0.500} & 0.371 & 0.657 & 0.800 & 0.800 & 0.771 & 0.800 & 0.800 & 0.800 & 0.800 & 0.771 \\
& pancreatic lesion resectability & \textcolor{gray}{0.333} & 0.379 & 0.483 & 0.483 & 0.483 & 0.483 & 0.414 & 0.483 & 0.483 & 0.483 & 0.483 \\
& pancreatic steatosis & \textcolor{gray}{0.500} & 0.526 & 0.526 & 0.513 & 0.513 & 0.579 & 0.526 & 0.526 & 0.526 & 0.526 & 0.658 \\
& pancreatic tumor staging & \textcolor{gray}{0.250} & 0.216 & 0.351 & 0.243 & 0.189 & 0.324 & 0.216 & 0.216 & 0.297 & 0.135 & 0.432 \\
\multicolumn{2}{c|}{\textbf{Average}} & \textcolor{gray}{0.403} & 0.446 & 0.557 & 0.560 & 0.555 & \textbf{0.584} & 0.550 & 0.562 & 0.561 & 0.549 & \textbf{0.629} \\
\midrule
\multicolumn{2}{c|}{\textbf{Total Average}} & \textcolor{gray}{0.369} & 0.440 & 0.626 & 0.632 & 0.628 & \textbf{0.662} & 0.478 & 0.489 & 0.493 & 0.497 & \textbf{0.555} \\
\bottomrule
\end{tabular}
\label{tab:vqa_full_comparison}
\end{table*}

\paragraph{2. Diagnostic competencies exhibit uneven model readiness.}
We primarily base this analysis on multi-choice accuracy, which is more stable and easier to interpret than free-text outputs.

\textit{Measurement} tasks are the most tractable, with all models significantly outperforming random-guess. This likely stems from the relatively large size and high signal-to-noise ratio of the anatomical targets (\eg organs or large lesions). Reasoning subtypes that involve volume aggregation or enlargement show similar trends, indicating that current VLMs can handle coarse quantification.

In contrast, \textit{recognition} tasks expose fundamental limitations. While accuracy may exceed 60\%, closer inspection reveals poor performance: most models default to majority-class answers, reflecting strong language priors and insufficient adaptation to subtle visual cues. RadFM, which is fully fine-tuned, is the only model that reliably escapes this bias; LoRA-based models fail to adjust generation tendencies.

\textit{Visual reasoning} tasks, which require combining recognition, localization, and measurement, reveal emerging but inconsistent capabilities. Models perform well on multi-step tasks like \textit{largest lesion diameter} or \textit{tumor-organ HU difference}, but struggle on fine-grained spatial subtypes like \textit{kidney volume comparison}, suggesting difficulty in bilateral reasoning and precise localization.

\textit{Medical reasoning} remains the most challenging category. These tasks require integrating imaging findings with domain knowledge not explicitly seen during training. RadFM again leads, likely benefiting from a larger language backbone and richer pretraining corpus. This points to the need for either diagnostic logic supervision or scaled multimodal instruction tuning.

Overall, while modern VLMs demonstrate promise in basic and recognition-heavy tasks, their applicability to real-world diagnostics is currently limited by weak visual signal, unreliable numeracy, and shallow reasoning chains.

\paragraph{3. Language model design influences VQA performance.}

\textbf{Scaling pretraining data enhances generalization.}  
RadFM achieves top performance across all tasks, particularly in recognition (multi-choice: 0.789; free-text: 0.812) and medical reasoning (free-text: 0.629). We attribute this to its large-scale pretraining (16M 2D+3D pairs), a 13B LLaMA2 decoder, and full model fine-tuning—highlighting the importance of both data scale and parameter updating. Future work should explore the tradeoff between tuning granularity and downstream adaptation.

\textbf{LLM size alone is not decisive.}  
Despite a smaller parameter count, M3D with Phi-3-4B performs slightly better than its LLaMA2-7B variant on visual (0.627 vs. 0.616) and medical reasoning (0.560 vs. 0.557). This suggests that under fixed vision modules, model size offers limited gains; architectural choice and pretraining strategy may matter more than scale alone~\citep{yousri2023big}.

\paragraph{4. Vision module choices significantly affect performance.}

\textbf{Vision encoder and projector design are critical.}  
Merlin adopts a 3D ResNet with a single global token projected via a linear layer, resulting in inferior performance. In contrast, RadFM, M3D, and CT-CHAT use ViT-style 3D encoders that produce token sequences, enabling richer spatial reasoning through attention. Token-level granularity appears essential for capturing complex volumetric patterns.

\textbf{Input spacing and resizing methods show weak correlation with performance.}  
RadFM and M3D resize raw CT volumes directly, whereas Merlin and CT-CHAT resample spacing and crop or pad to target dimensions. In theory, spacing inconsistency may degrade volume-sensitive measurements, yet we observe that direct resizing does not hurt performance on tasks such as \textit{organ volume measurement}, \textit{lesion volume measurement}, and \textit{kidney volume comparison}. Similarly, CT-CHAT receives the largest input size ([300, 300, 600]) but underperforms across most reasoning tasks. Merlin processes [224, 224, 160] volumes but yields even lower overall accuracy. These results indicate that \textbf{larger input resolution or spacing alignment alone is insufficient to ensure better diagnostic performance}.

\subsection{Impact of Measurement and Recognition on Reasoning Tasks.}

\begin{figure}
    \centering
    \includegraphics[width=1.0\linewidth]{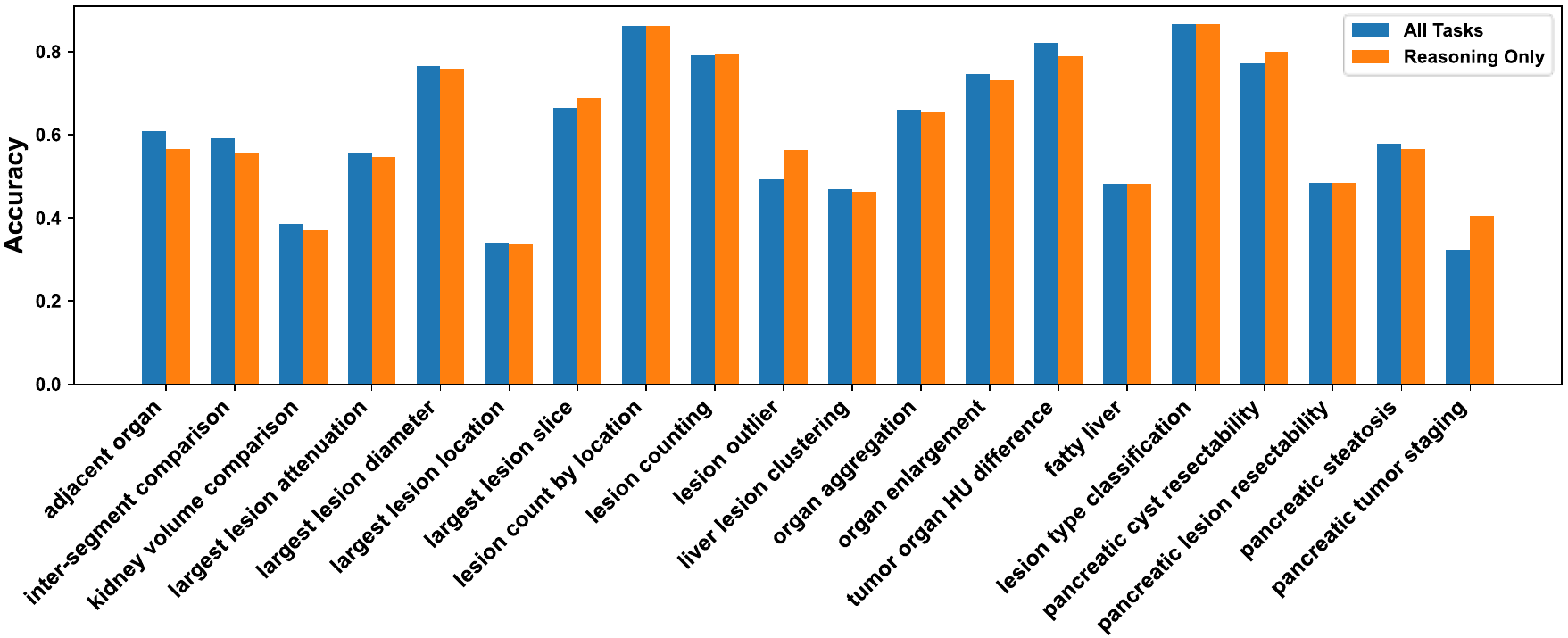}
    \caption{The RadFM accuracy of reasoning tasks with or without measurement/recognition tasks.}
    \label{fig:acc_reasoning_only}
\end{figure}

We train RadFM without measurement and recognition tasks to see whether there is a crucial impact of basic tasks on higher-level tasks. The relatively small performance gap in Fig. \ref{fig:acc_reasoning_only} suggests that RadFM already generalizes reasonably well to reasoning tasks, regardless of whether measurement/recognition is explicitly seen during training. We hypothesize the main reason is that RadFM was pre-trained on large-scale 2D/3D image-text data, including structured reports, which may implicitly cover recognition and measurement concepts. But we still find that in several subtypes like \textit{inter-segment comparison} and \textit{tumor organ HU difference}, all-tasks training brings a notable benefit. These subtypes may heavily rely on the explicit annotation of liver subsegments and HU values in the DeepTumorVQA dataset.

\begin{figure}
\centering
\includegraphics[width=0.8\linewidth]{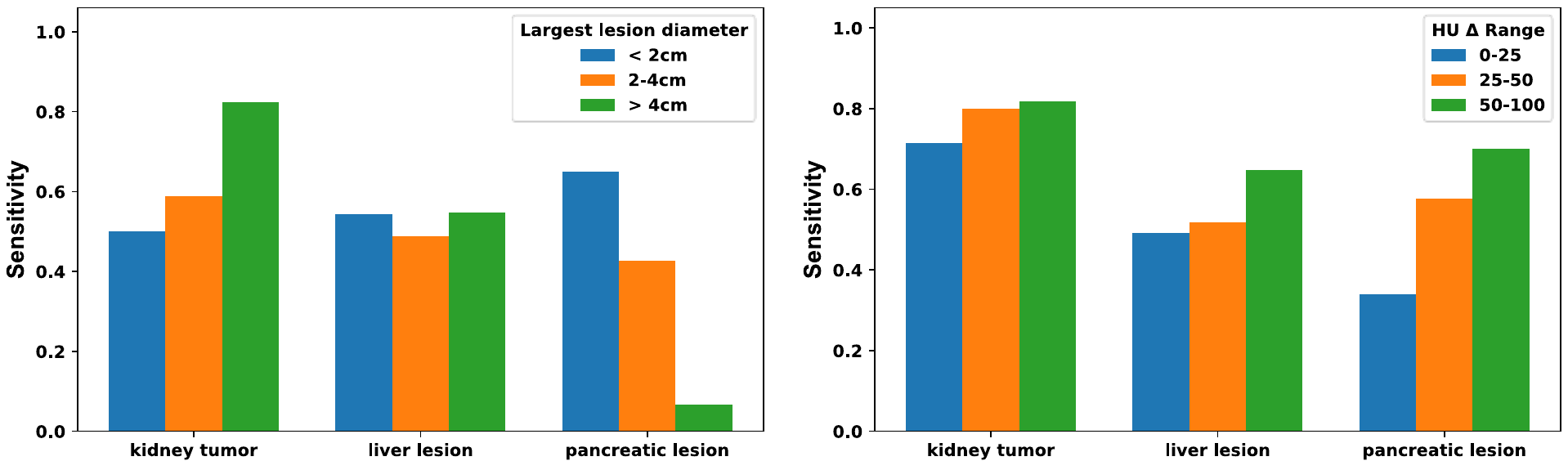}
\caption{
Lesion recognition sensitivity of RadFM under different lesion sizes (left) and HU contrast ranges (right).  
\textbf{Left:} Sensitivity increases with size only for kidney tumors, while liver and pancreatic lesions show no consistent trend.  
\textbf{Right:} Higher HU contrast leads to higher sensitivity across all lesion types, indicating that intensity-based features significantly affect detection performance.
}
\label{fig:lesion_sensitivity_analysis}
\end{figure}

\subsection{Effect of Lesion Size and HU Contrast on Recognition Sensitivity.}

To better understand the factors that influence lesion recognition performance, we analyze RadFM's recognition sensitivity across different lesion sizes and HU contrasts. 

Figure~\ref{fig:lesion_sensitivity_analysis} (left) shows sensitivity grouped by lesion diameter (<2cm, 2–4cm, >4cm). For kidney tumors, sensitivity increases with size. However, liver and pancreatic lesions do not follow this trend; in particular, sensitivity for large pancreatic lesions decreases. We hypothesize that this may stem from anatomical complexity obscuring large lesions, annotation imbalance, or model reliance on contextual rather than absolute size cues.
In contrast, Figure~\ref{fig:lesion_sensitivity_analysis} (right) shows a consistent increase in sensitivity with larger lesion-to-organ HU differences (0–25, 25–50, 50–100). This trend holds across all lesion types and suggests that stronger intensity contrast enhances boundary detectability, making HU difference a more reliable predictor of VLM sensitivity than physical size.


\subsection{Improving Lesion Recognition via Segmentation-based Preprocessing.}


\begin{table}[t]
\centering
\scriptsize
\caption{Lesion recognition sensitivity, specificity, and accuracy (\%) for three organs across nnUNet (oracle), existing VLMs, and our proposed nnM3D that uses nnUnet for organ localization.}
\label{tab:lesion_recognition_table}
\begin{tabular}{l|ccc|ccc|ccc}
\toprule
\multirow{2}{*}{\textbf{Model}} & \multicolumn{3}{c|}{\textbf{Liver Lesion}} & \multicolumn{3}{c|}{\textbf{Kidney Tumor}} & \multicolumn{3}{c}{\textbf{Pancreatic Lesion}} \\
& Sens. & Spec. & Acc. & Sens. & Spec. & Acc. & Sens. & Spec. & Acc.  \\
\midrule
nnUNet (oracle) & 86.2 & 73.4 & 81.7 & 96.3 & 78.3 & 87.7 & 80.0 & 76.6 & 78.9 \\
\hline
RadFM & 53.0 & 78.6 & 65.2 & 75.2 & 98.6 & 88.6 & 40.3 & 97.0 & 81.0 \\
M3D-Phi3 & 90.8 & 8.7 & 51.7 & 0.0 & 100.0 & 57.4 & 0.0 & 100.0 & 71.8 \\
M3D-LLaMA2 & 100.0 & 0.0 & 52.4 & 0.0 & 100.0 & 57.4 & 0.0 & 100.0 & 71.8 \\
Merlin & 52.7 & 52.0 & 52.4 & 50.2 & 51.2 & 50.7 & 48.9 & 51.6 & 50.8 \\
CT-CHAT & 100.0 & 0.0 & 52.4 & 0.0 & 100.0 & 57.4 & 0.0 & 100.0 & 71.8 \\
\rowcolor{gray!10}
nnM3D-Phi3 & 63.7 & 62.0 & 62.9 & 79.1 & 95.7 & 88.6 & 2.6 & 98.2 & 71.3 \\
\rowcolor{gray!10}
nnM3D-LLaMA2 & 67.6 & 58.6 & 66.3 & 80.9 & 95.3 & 89.2 & 35.1 & 91.6 & 75.7 \\
\bottomrule
\end{tabular}
\end{table}

Despite their success on general VQA tasks, current VLMs exhibit substantial failures in lesion recognition, especially for small tumors. As shown in Table~\ref{tab:lesion_recognition_table}, several models (\eg M3D-LLaMA2, M3D-Phi3, CT-CHAT) collapse into predicting the dominant class across all samples, leading to imbalanced sensitivity and specificity. This indicates that without explicit spatial localization, VLMs fail to attend to subtle lesion signals in raw 3D volumes.

To address this, we propose a simple yet effective strategy that crops the input images around target organs through nnUNet~\citep{isensee2021nnu} anatomical localization. This approach reduces the noisy information of irrelevant regions and zooms in on target organs. We denote the resulting models as \textbf{nnVLM} variants.

Our experiments in Table~\ref{tab:lesion_recognition_table} show that nnM3D achieves substantial gains in lesion recognition across all three organs. For instance, nnM3D-LLaMA2 improves kidney tumor sensitivity from 0\% to 80.9\%, surpassing even RadFM in this case. These results highlight the importance of anatomical context in vision-language learning, and suggest that simple localization priors can serve as effective alternatives to full voxel-level supervision.

\begin{figure}
\centering
\includegraphics[width=0.8\linewidth]{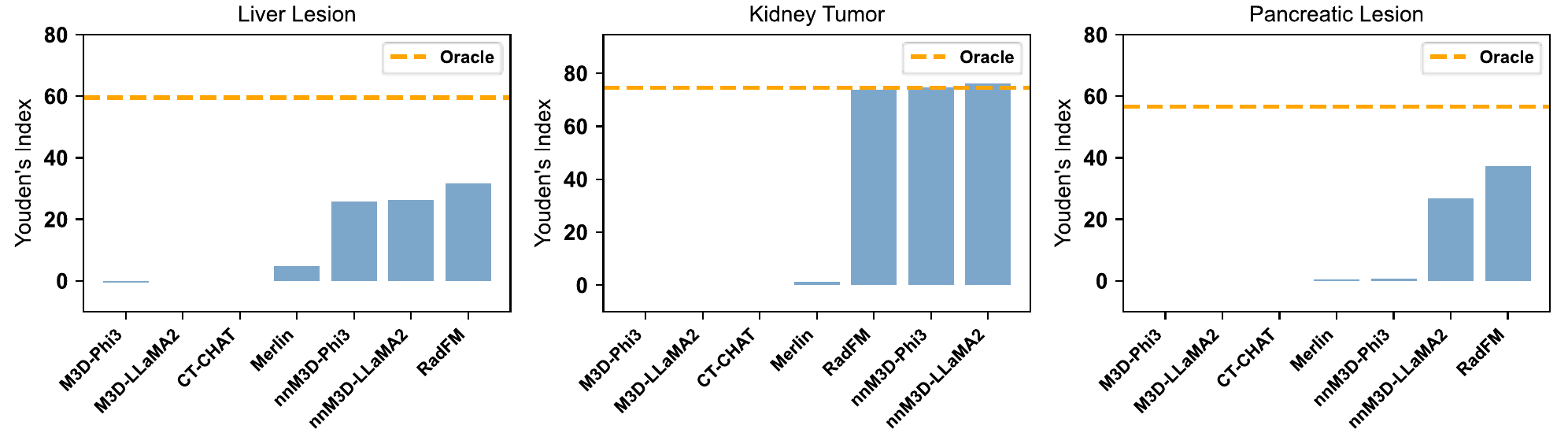}
\caption{Comparison of Youden's Index (sensitivity + specificity - 1) of VLMs and the oracle.}
\label{fig:youden_index}
\vspace{-0.3cm}
\end{figure}

Figure~\ref{fig:youden_index} further visualizes model-level performance using Youden's index. While liver and pancreas remain challenging, multiple VLMs approach or match the oracle's performance on kidney tumors. This suggests that with targeted preprocessing, medical VLMs may close the gap with segmentation-based recognition methods. We expect our benchmark can witness VLMs' improvement, getting closer or even surpassing the segmentation methods.



\section{Discussion and Conclusion}
\label{sect:discussion}

\textbf{Are 3D medical VLMs precise and intelligent enough for clinical diagnosis?} 
This work takes a step toward answering this question by introducing \textbf{DeepTumorVQA}, the first large-scale VQA benchmark focused on 3D clinical diagnosis that enables not only quantitative evaluation but also tracing of model failures. Through extensive evaluation, our dataset reveals that while VLMs exhibit emerging precision in basic measurement and recognition (even approaching segmentation models), their overall intelligence remains far from meeting clinical requirements, especially in medical reasoning tasks. Through careful inspection, we reveal the impact of basic tasks on the reasoning task, and also analyze the difficulty of lesion recognition \wrt both lesion size and HU contrast.

\textbf{The dataset exposes critical differences in 3D medical VLMs.}  
First, visual architectures matter: our experiments show that ViT-based 3D encoders significantly outperform single-token CNN backbones in tasks requiring spatial reasoning or multi-lesion aggregation. Second, language decoder scale alone does not guarantee improved performance; rather, large-scale pretraining and full-tuning strategies—exemplified by RadFM—yield more consistent gains across tasks. Third, our proposed organ-specific preprocessing pipeline demonstrates that vision models with anatomical priors significantly improve lesion detection by spatial localization. 

\textbf{Limitations.} The dataset construction process relies heavily on precise organ and lesion segmentation to generate structured metadata and QA pairs. However, due to the inherent variability in radiologist expertise and the ambiguity of certain CT appearances (\eg low-contrast lesions or anatomical variants), the imperfect segmentation quality may introduce noise into downstream QA pairs. The dataset is intended as a research benchmark, not for clinical deployment or decision-making that may cause risks for false reassurance or missed diagnoses. Additionally, while our experiments provide insightful comparisons across vision-language model architectures and training regimes, the conclusions would benefit from more controlled ablation studies to isolate variables systematically.

\textbf{Conclusion.}  
DeepTumorVQA fills a critical gap in the evaluation of medical VLMs. It serves both as a diagnostic tool and as a development benchmark. We will hold recurring challenges to support the community in building safer, more explainable, and ultimately clinically useful multimodal systems.

\begin{ack}
This work was supported by the Lustgarten Foundation for Pancreatic Cancer Research and the Patrick J. McGovern Foundation Award.
\end{ack}

\bibliographystyle{plainnat}
\bibliography{TumorVQA}

\newpage
\appendix

\section{Metadata and Structured Description Generation}
\label{sect:appendix_A}

To support systematic question generation and fine-grained model evaluation, we construct a rich set of structured metadata for each CT volume using paired organ and lesion segmentation masks. This section describes how we derive the metadata fields and generate a radiology-style structured description for each case.

\subsection{Metadata Extraction from Segmentation Masks}

Given a CT volume and corresponding 3D segmentation masks for organs and lesions, we extract anatomical and lesion-level statistics through the following steps:

\begin{itemize}
    \item \textbf{Resampling and alignment:} All masks are resampled to the same voxel spacing as the CT image. We ensure alignment across volumes and segmentations to preserve geometric correctness.
    \item \textbf{Volume and size statistics:} For each organ and its lesions (\eg liver, kidney, pancreas), we compute total organ volume, total lesion volume, and the number of lesion instances.
    \item \textbf{Largest lesion analysis:} We extract the size (diameter), location (\eg liver segment or organ side), and mean attenuation (HU value) of the largest lesion per organ and subtype (tumor, cyst, or unspecified lesion).
    \item \textbf{Enhancement type classification:} Using the HU value difference between lesions and organ parenchyma, we classify lesion attenuation into three categories: \textit{hyperattenuating}, \textit{isoattenuating}, and \textit{hypoattenuating}.
    \item \textbf{Clinical staging:} For pancreas tumors, we approximate T-stage (T1–T4) based on existing staging protocols.
    \item \textbf{Demographic and acquisition metadata:} Patient age, sex, contrast phase, and scanner type are retrieved from DICOM headers or accompanying metadata files.
\end{itemize}

The final metadata table includes over 70 structured attributes per scan, such as:
\textit{liver lesion count, largest kidney tumor diameter (cm), pancreatic tumor attenuation, spleen volume, organ HU values, lesion location,} and more. This table enables compositional and interpretable question generation across a wide range of diagnostic concepts.

\subsection{Structured Report-style Description}

In addition to the tabular metadata, we generate a structured textual description in radiology report style for each scan following \citep{bassi2025radgpt}. This free-text summary provides high-resolution lesion-level information and mimics real radiological narratives. Each description includes:

\begin{itemize}
    \item A global summary per organ (\eg volume, mean HU).
    \item Instance-level lesion summaries: lesion size, volume, location (\eg liver segment, pancreas head/body/tail), slice number, attenuation classification.
    \item Aggregated lesion counts and total tumor/cyst volumes.
    \item Impression statement summarizing major findings, such as:  
    \textit{“Multiple (25) hypoattenuating liver masses. Largest one (segment 2) measures 3.2 x 1.7 cm. Total volume of all liver masses: 19.4 cm\textsuperscript{3}.”}
\end{itemize}

An example of the structured description is shown in the following. 

\begin{tcolorbox}[title=Example: Radiology-style Structured Report, colback=gray!5, colframe=gray!50!black, fonttitle=\bfseries, breakable]
CT Arterial Phase\\

FINDINGS:\\

Liver:
Normal size (volume: 1293.7 cm³).
Mean HU value: 111.3 ± 17.4.\\

Liver lesions:
Liver lesion 1:
Location: hepatic segment 2.
Size: 3.2 x 1.7 cm (image 174). Volume: 8.1 cm³.
Enhancement relative to liver: Hypoattenuating (HU value is 9.6 ± 19.8).\\

Liver lesion 2:
Location: hepatic segment 8.
Size: 2.5 x 2.0 cm (image 178). Volume: 4.9 cm³.
Enhancement relative to liver: Hypoattenuating (HU value is 36.3 ± 31.3).\\

... [truncated for brevity]\\

Liver lesion 24: 
Location: hepatic segment 5.
Size: 0.5 x 0.4 cm (image 156). Volume: 0.1 cm³.
Enhancement relative to liver: Hypoattenuating (HU value is 97.2 ± 16.9).\\

Liver lesion 25: 
Location: hepatic segment 4.
Size: 0.4 x 0.3 cm (image 156). Volume: 0.0 cm³.
Enhancement relative to liver: Hypoattenuating (HU value is 71.3 ± 20.0).\\

Pancreas: 
Normal size (volume: 80.3 cm³).
Mean HU value: 104.8 ± 28.5.\\

Kidney: 
Normal size (right kidney volume: 166.6 cm³; left kidney volume: 156.6 cm³; total kidney volume: 323.2 cm³). Mean HU value: 127.4 ± 52.8.\\

Spleen: 
Normal size (volume: 135.1 cm³).
Mean HU value: 124.7 ± 34.8.\\

IMPRESSION:
Multiple (25) hypoattenuating liver masses. Largest one (hepatic segment 2) measures 3.2 x 1.7 cm.
Total volume of all liver masses: 19.4 cm³.
\end{tcolorbox}

These descriptions support reasoning question generation (\eg ``How are the liver lesions distributed across hepatic segments") and provide explainable context for model output interpretation.

\section{Task Definitions and Generation Logic}
\label{sect:appendix_B}

To support a systematic and diverse evaluation of vision-language models in 3D tumor-centric diagnosis, DeepTumorVQA includes 29 question subtypes spanning four diagnostic categories: \textit{Measurement}, \textit{Recognition}, \textit{Visual Reasoning}, and \textit{Medical Reasoning}. 

Each question subtype corresponds to a well-defined clinical concept (e.g., organ size, lesion count, resectability), and is generated through a rule-based or metadata-driven functional program. These question types are designed to reflect increasing levels of diagnostic complexity, ranging from direct retrieval to multi-step inference.

Table~\ref{tab:task_definitions} summarizes all subtypes, their task type, the logic used for answer generation, and an example QA pair. This structured taxonomy enables reproducible benchmarking and compositional analysis of VLM performance across clinical tasks.

\begin{center}
\setlength{\tabcolsep}{4pt}  
\renewcommand{\arraystretch}{1.1}  
\begin{longtable}{
>{\scriptsize}p{2.5cm} >{\scriptsize}p{3.2cm} 
>{\scriptsize}p{3.5cm} >{\scriptsize}p{3.5cm}
}
\caption{Summary of task types, subtypes, generation logic, and example question-answer pairs in \textbf{DeepTumorVQA}. Tasks are organized by their diagnostic intent: measurement, recognition, visual reasoning, and medical reasoning.}
\label{tab:task_definitions} \\
\toprule
\textbf{Task Type} & \textbf{Subtype} & \textbf{Definition / Generation Logic} & \textbf{Example QA Pair} \\
\midrule
\endfirsthead

\multicolumn{4}{l}{\textit{(continued from previous page)}} \\
\toprule
\textbf{Task Type} & \textbf{Subtype} & \textbf{Definition / Generation Logic} & \textbf{Example QA Pair} \\
\midrule
\endhead

\bottomrule
\endfoot

Measurement & organ volume measurement & Quantify organ size from metadata using volume. & Q: What is the liver volume? A: 1293.7 cm³ \\
Measurement & organ HU measurement & Extract organ mean HU value using regex from report. & Q: What is the mean HU of the pancreas? A: 104.8 \\
Measurement & lesion volume measurement & Sum total lesion volume from metadata (per lesion type and organ). & Q: What is the total tumor volume in the right kidney? A: 17.5 cm³ \\

Recognition & liver lesion existence & Check presence of any lesion in liver by total volume > 0. & Q: Is there any lesion in the liver? A: Yes \\
Recognition & pancreatic lesion existence & Check presence of any lesion in pancreas. & Q: Does the pancreas have any lesions? A: No \\
Recognition & kidney lesion existence & Check presence of non-specific kidney lesions. & Do we have evidence of any lesions within the kidney? A: No \\
Recognition & kidney cyst existence & Check presence of cyst in kidney. & Is there at least one cyst in the kidney? A: No \\
Recognition & kidney tumor existence & Check presence of tumor in kidney. & Would the kidney be described as having tumors? A: Yes \\
Recognition & colon lesion existence & Check presence of colon lesions. & Is the colon affected by any lesions? A: No \\

Visual Reasoning & lesion counting & Count lesion instances by type and organ. & Q: How many cysts are there in the liver? A: 3 \\
Visual Reasoning & largest lesion diameter & Use metadata field for largest lesion diameter. & Q: What is the diameter of the largest tumor in the pancreas? A: 2.5 cm \\
Visual Reasoning & largest lesion location & Read lesion location label (e.g. segment 1–8 or left/right). & Q: Where is the largest liver lesion located? A: Segment 2 \\
Visual Reasoning & largest lesion attenuation & Classify lesion HU vs. background as hypo/iso/hyper. & Q: Is the largest liver cyst hypoattenuating? A: Yes \\
Visual Reasoning & kidney volume comparison & Compare left/right kidney volumes and discretize into 3 options. & Q: Which kidney is larger? A: Left kidney \\
Visual Reasoning & organ aggregation & Sum two organs' volumes. & Q: What is the combined volume of liver and spleen? A: 1428.3 cm³ \\
Visual Reasoning & tumor organ HU difference & Compute absolute HU diff between lesion and corresponding organ. & Q: What is the HU difference between kidney tumor and kidney? A: 32.4 \\
Visual Reasoning & largest lesion slice & Locate axial slice with max lesion size and normalize by depth. & Q: On which slice is the largest liver lesion found? A: Slice 174 \\
Visual Reasoning & lesion outlier & If largest lesion is >3× volume of second largest → outlier. & Q: Is the largest lesion 3× larger than the second largest? A: No \\
Visual Reasoning & lesion count by location & Extract per-segment or sub-region lesion counts from report. & Q: How many liver lesions are in segment 8? A: 5 \\
Visual Reasoning & inter-segment comparison & Compare lesion counts between two liver segments. & Q: Which segment has more lesions: segment 2 or 4? A: Segment 2 \\
Visual Reasoning & adjacent organ & Extract from text: reported adjacent organ names for largest lesion. & Q: Which organ is adjacent to the largest liver lesion? A: Stomach \\
Visual Reasoning & organ enlargement & Use `enlarged' keyword from report per organ. & Q: Is the pancreas enlarged? A: No \\
Visual Reasoning & liver lesion clustering & If > 3 lesions within 3 adjacent segments, mark as `clustered'. & Q: Are liver lesions clustered in adjacent segments? A: Yes \\

Medical Reasoning & pancreatic tumor staging & Use labeled T-stage for pancreatic tumor. & Q: What is the stage of the pancreatic tumor? A: T2 \\
Medical Reasoning & fatty liver & Use liver/spleen HU ratio and liver HU to classify steatosis severity. & Q: Does the liver show fatty infiltration? A: Yes \\
Medical Reasoning & pancreatic steatosis & Use pancreas/spleen HU ratio to assess steatosis (<0.7 = Yes). & Q: Does the pancreas show steatosis? A: No \\
Medical Reasoning & pancreatic cyst resectability & Binary classification: cyst volume > 3.0 cm³ → resection. & Q: Is the pancreatic cyst resectable? A: Yes \\
Medical Reasoning & lesion type classification & If largest kidney lesion HU > threshold → tumor else cyst. & Q: Is the kidney lesion a cyst or tumor? A: Tumor \\
Medical Reasoning & pancreatic lesion resectability & Use largest lesion's report-tagged resectability field. & Q: Can the pancreatic lesion be surgically resected? A: No \\

\end{longtable}
\end{center}

\section{Training details for VLMs}
\label{sect:appendix_C}

We provide detailed training configurations for the four benchmarked vision-language models (VLMs) evaluated in this work: RadFM, M3D (with both LLaMA2 and Phi-3 decoders), Merlin, and CT-CHAT. To ensure a fair comparison, all models are trained using their official open-source codebases and adapted to the DeepTumorVQA dataset with minimal changes to architecture or optimization logic.

Table~\ref{tab:hyperparameters} summarizes key hyperparameters and compute resource settings. All models are trained with AdamW optimizer and cosine learning rate scheduling. Mixed-precision training is enabled using either FP16 or BF16, depending on framework compatibility. For large models such as RadFM and M3D, gradient accumulation is used to simulate larger batch sizes, with 4 GPUs and 16 CPU workers for data loading.

Notably, due to high memory requirements, Merlin is trained with a batch size of 1 and gradient accumulation of 8, while CT-CHAT benefits from a higher per-device batch size due to its lighter vision backbone. Training for all models is conducted for approximately 48 hours using commodity GPU clusters (NVIDIA A5000, A6000, and A100 as indicated).

\begin{table*}[ht]
\centering
\scriptsize
\caption{Model training hyperparameters and compute resource for four benchmarked VLMs.}
\begin{tabular}{lcccc}
\toprule
\textbf{Item} & \textbf{RadFM} & \textbf{M3D (LLaMA2 / Phi-3)} & \textbf{Merlin} & \textbf{CT-CHAT} \\
\midrule
Learning rate & 5e-6 & 5e-5 & 1e-4 & 2e-5 \\
Optimizer & AdamW (8-bit) & AdamW & AdamW & AdamW \\
Auto mixed precision & FP16 & BF16 & BF16 & FP16 \\
Per device batch size & 4 (model parallel) & 1 & 1 & 32 \\
Gradient accumulation steps & 8 & 8 & 8 & 1 \\
Learning rate scheduler & Cosine & Cosine & Cosine & Cosine \\
Warmup ratio & 0 & 0.03 & 0.03 & 0.03 \\
Training iterations & 25k & 33k & 25k & 3 epochs \\
CPU workers & 16 & 16 & 16 & 128 \\
GPU hardware & 4×A5000 24GB & 4×A6000 48GB & 4×A5000 24GB & 4×A100 80GB \\
RAM & 128GB  & 128GB  & 256GB  & 1024GB \\
Compute time & 48 hours & 48 hours & 48 hours & 48 hours \\
\bottomrule
\end{tabular}
\label{tab:hyperparameters}
\end{table*}

\section{Accuracy Breakdown across Demographic and Imaging Factors}
\label{sect:appendix_D}

To explore whether vision-language model (VLM) performance varies across patient or scan-related subgroups, we stratify question-answering accuracy by four categorical factors extracted from metadata: sex, age group, CT scanner manufacturer, and contrast phase. Accuracy is reported per question category: \textit{measurement}, \textit{recognition}, \textit{visual reasoning}, and \textit{medical reasoning}.

\begin{figure}[t]
\centering
\includegraphics[width=1.0\textwidth]{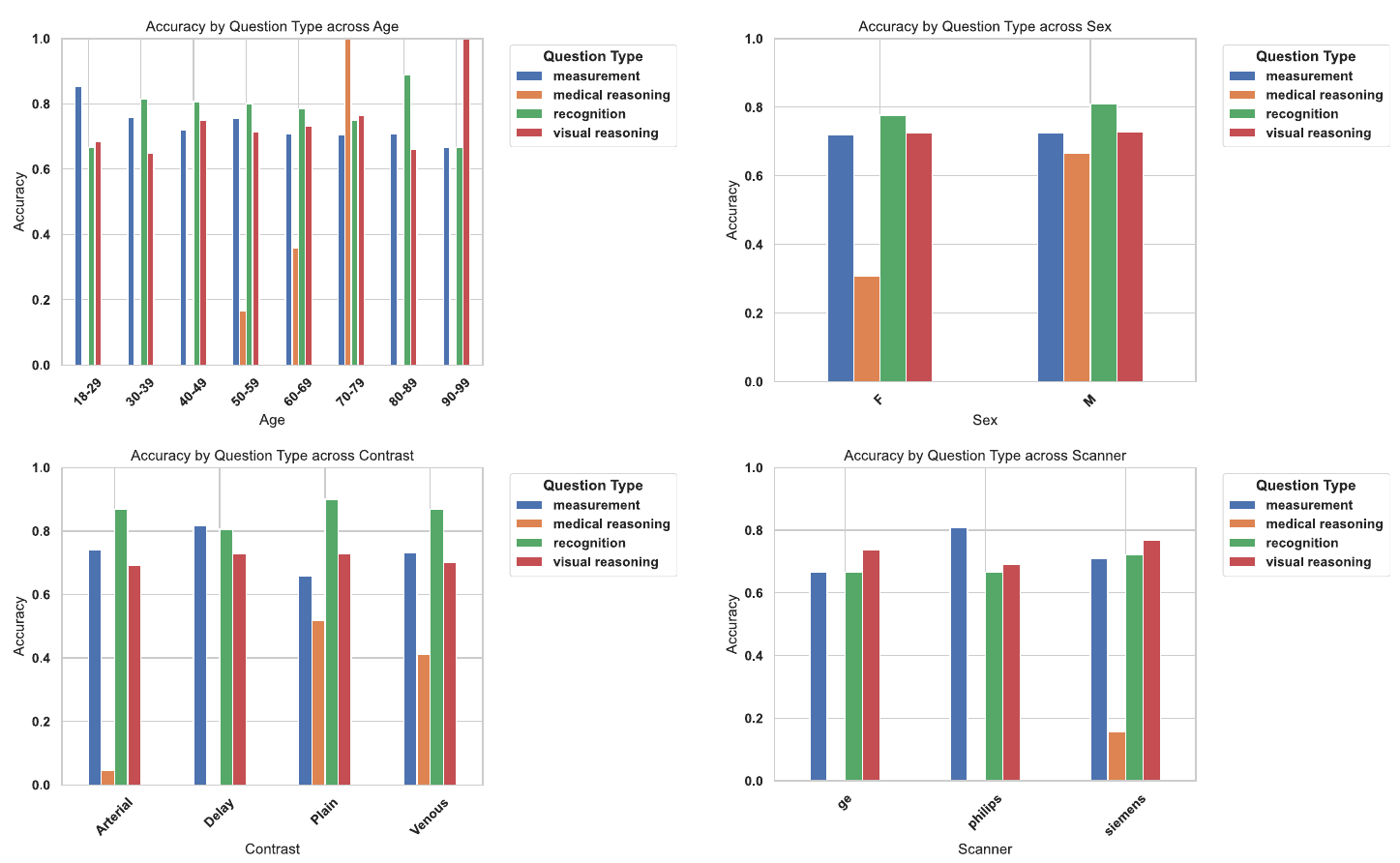}
\caption{Accuracy by question type across sex.}
\label{fig:acc_by_metadata}
\end{figure}

\paragraph{Age.} Figure~\ref{fig:acc_by_metadata} (upper left) shows that recognition and measurement tasks remain stable across most age groups, while medical reasoning accuracy is more volatile. Notably, large drops are observed in 60--69 and 90--99 bins for medical reasoning, which may reflect either smaller sample size or increased scan complexity (\eg, multi-lesion, ambiguous enhancement). This underscores the importance of stratified evaluation in medical datasets.

\paragraph{Sex.} As shown in Figure~\ref{fig:acc_by_metadata} (upper right), overall performance is similar across female (F) and male (M) cohorts, with no substantial gap in any task type. Recognition is the strongest category in both groups. A slight improvement in medical reasoning is observed in males, possibly due to distributional biases in training samples (\eg, sex imbalance in pancreas/uterus-related cases).

\paragraph{Contrast Phase.} As shown in Figure~\ref{fig:acc_by_metadata} (lower left), recognition accuracy is high and stable across all contrast phases (arterial, delay, plain, venous), suggesting robustness of perception to intensity changes. However, medical reasoning suffers in the arterial and venous phases, likely due to poor organ-lesion contrast or increased noise in attenuation-based reasoning (\eg, fatty liver, lesion enhancement).

\paragraph{Scanner Manufacturer.} In Figure~\ref{fig:acc_by_metadata} (lower right), all three vendors (GE, Philips, Siemens) show consistent performance on measurement and recognition tasks. However, a sharp drop in medical reasoning accuracy is observed for Siemens, potentially due to domain shift in intensity values or HU calibration differences, which may affect reasoning modules trained on scanner-agnostic data.

These results suggest that while modern VLMs can generalize well across standard factors like sex and age, their medical reasoning performance may be more sensitive to acquisition protocol and scanner variation. Future work should incorporate domain adaptation or uncertainty modeling to ensure reliability across subpopulations.

\end{document}